\documentclass[11pt]{article}

\usepackage{caption}
\usepackage{url}
\usepackage{subfigure}
\usepackage[T1]{fontenc}
\usepackage[utf8]{inputenc}
\usepackage{authblk}
\usepackage{color}
\usepackage{amsmath, amssymb, latexsym}
\usepackage{psfrag,epsfig,amsfonts,amsmath,latexsym,amsthm,amssymb,amscd,url }
\usepackage[colorlinks=true,linkcolor=blue,citecolor=blue, linktocpage=true]{hyperref}

\renewcommand{\phi}{\varphi}

\newcommand{\BE}{\begin{equation}}
\newcommand{\EE}{\end{equation}}
\newcommand{\BEN}{\begin{equation*}}
\newcommand{\EEN}{\end{equation*}}
\newcommand{\BAL}{\begin{align}}
\newcommand{\EAL}{\end{align}}
\newcommand{\BAN}{\begin{align*}}

\begin{document}

\title{Learning the temporal evolution of multivariate densities\\ via normalizing flows}

\author[a]{Yubin Lu}
\author[b,c]{Romit Maulik \thanks{Corresponding author: rmaulik@anl.gov}}
\author[a]{Ting Gao}
\author[d]{Felix Dietrich}
\author[e]{Ioannis G. Kevrekidis}
\author[c]{Jinqiao Duan}
\affil[a]{School of Mathematics and Statistics \& Center for Mathematical Sciences, Huazhong University of Science and Technology, Wuhan 430074, China}
\affil[b]{Argonne Leadership Computing Facility, Argonne National Laboratory, Lemont, IL 60439}
\affil[c]{Department of Applied Mathematics, College of Computing, Illinois Institute of Technology, Chicago, IL 60616, USA}
\affil[d]{Department of Informatics, Technical University of Munich, Boltzmannstr. 3, 85748 Garching b. Munich, Germany}
\affil[e]{Departments of Applied Mathematics and Statistics \& Chemical and Biomolecular Engineering, Johns Hopkins University, Baltimore, Maryland 21211, USA}

\renewcommand*{\Affilfont}{\small\it}
\renewcommand\Authands{ and }
\date{\today}
\maketitle

 \begin{abstract}

In this work, we propose a method to learn multivariate probability distributions using sample path data from stochastic differential equations. Specifically, we consider temporally evolving probability distributions (e.g., those produced by integrating local or nonlocal Fokker-Planck equations). We analyze this evolution through machine learning assisted construction of a time-dependent mapping that takes a reference distribution (say, a Gaussian) to each and every instance of our evolving distribution. If the reference distribution is the initial condition of a Fokker-Planck equation, what we learn is the time-T map of the corresponding solution. Specifically, the learned map is a multivariate normalizing flow that deforms the support of the reference density to the support of each and every density snapshot in time. We demonstrate that this approach can approximate probability density function evolutions in time from observed sampled data for systems driven by both Brownian and L\'evy noise. We present examples with two- and three-dimensional, uni- and multimodal distributions to validate the method.

\end{abstract}

\bigskip\par

%\tableofcontents

%%%%%%%%%%%%%%%%%%%%%%%% Body
%(for instance drift, diffusion, quantities of interest such as mean residence times, escape probabilities, etc)

\textsl{Stochastic differential equations (SDEs) have traditionally been used to create informative deterministic models for features of stochastic dynamics. SDEs are frequently accompanied by Fokker-Planck partial differential equations (PDEs) which are used to calculate quantities of interest such as the mean residence time, escape probabilities and transition probabilities for the underlying stochastic processes. Recently, data-driven methods for learning the complex dynamics of stochastic processes have attracted increasing attention. Specifically,
there has been a burst of algorithm development for the data-driven identification of stochastic systems through accurate and robust probability density estimation. In this article, we devise a method to approximate time-varying high-dimensional and multimodal distributions (for e.g., given by the solutions of non-local Fokker-Planck equations for dynamical systems driven by Brownian and L\'evy noise) from sample path data. We validate the approach by investigating several two- and three-dimensional stochastic processes, where we compare predictions from our method with the true solutions of their respective Fokker-Planck equations.}

\section{Introduction}

Stochastic differential equations (SDEs) are often used to model complex systems in science and engineering under the influence of different forms of randomness, such as fluctuating forces or random boundary conditions. In this work, we are interested in learning the probability distributions for such systems driven by random perturbations. Typically, the Fokker-Planck equation (FPE) describes the probability density evolution of the solution for certain SDEs. These FPEs may be local or non-local depending on the nature of the noise that drives the SDE. With the help of the Fokker-Planck equations, we can subsequently estimate various quantities of interest, such as mean residence  times, escape probabilities, as well as the most probable transition pathways for extreme events, from time series data of biophysical, climate, or other complex systems with stochastic dynamics \cite{Duan2015}.

In recent times, data-driven analysis of complex stochastic dynamics has become widespread across various fields \cite{yang2020generative, weinan2017deep, beck2018solving, guler2019towards, tzen2019neural, jia2019neural, Kevrekidis1,Kevrekidis2,Kevrekidis3, Kevrekidis4}. Examples range from the identification of the reduced stochastic Burgers equations \cite{FeiLu}, to parameter estimation of stochastic differential equations using Gaussian processes \cite{GPs1,GPs2}, sparse identification of nonlinear dynamics  \cite{SINDy1,SINDy2}, Kramers–Moyal expansions for estimating densities from sample paths \cite{KM}, utilization of the Koopman operator formalism for analyzing complex dynamical systems \cite{DMD,EDMD,SKO1,SKO2,SKO3}, and implementing transfer operators for analyzing stochastic dynamics \cite{Dellnitz,Froyland,Klus1,Klus2,Metzner,Tantet,Thiede}, among others.

Recent data-driven research has made substantial progress in learning the dynamics of stochastic processes. For instance, \cite{LiuYang,Duvenaud1,Duvenaud2,jia2019neural,tzen2019neural,dietrich2021learning,Kevrekidis8, Kevrekidis9,Kevrekidis10, Kevrekidis5, Kevrekidis6, Kevrekidis7} parameterized the drift and diffusivity of stochastic differential equations as neural networks and trained these networks using sample path data. In \cite{yang2018physics}, generative adversarial networks (GANS) were used to learn SDEs by incorporating SDE knowledge into the loss function. In our previous works such as  \cite{XiaoliChen,DaiMinChaos,YangLi2020a,LuYB2020}, we have also considered the learning of stochastic differential equations driven by L\'evy noise and their corresponding non-local Fokker-Planck equations from data using neural networks, non-local Kramers–Moyal expansions, and Koopman operator analysis. A common theme of many of these studies is that \emph{a priori} knowledge of the nature of the SDE (state dependent diffusivity; Brownian noise; etc.) or the Fokker-Planck equations is required. We are inspired by the necessity to learn stochastic dynamics without explicit knowledge of the underlying system of equations. Consequently, we use modern generative machine learning techniques to learn bijective transformations between the support of reference and target densities using normalizing flows \cite{NFs1,NFs2,IAF,MAF,RealNVP} from sample data alone. In contrast with traditional uses of normalizing flows, we explicitly embed a dependence on physical time into the learnable transformations between supports/distributions. This is inspired by the preliminary studies in \cite{stNFs} where such a transformation was obtained for {\em scalar} stochastic processes. Here, we generalize to multidimensional scenarios. We also note that, in contrast to GANs, which only defines a stochastic procedure that directly generates data, the normalizing flow provides an explicit expression for the density by virtue of an analytical transformation from a known density function. Also, the normalizing flow does not prescribe an approximate density, like the variational autoencoder, for the purpose of learning.

The idea of a normalizing flow was introduced by Tabak and Vanden-Eijnden in the context of density estimation and sampling~\cite{tabak-2010}. A flow that is ``normalizing'' transforms a given distribution, generally into the standard normal, while its (invertible) Jacobian is traced and inverted continuously to obtain the inverse map. This results in the desired mapping (called ``a flow"), from the reference to the target density. Flows comprising the composition of neural network transformations were proposed later~\cite{NFs1}, with further recent progress on neural ODE-based mappings between supports of densities~\cite{Duvenaud1}. For a general review of current methods, see~\cite{Kobyzev-2019}. An earlier version of normalizing flows, coined ``optical flows" has been developed for image processing and alignment in computer vision~\cite{Horn-1981,Lucas-1981}. Very similar ideas are used in that domain, but applications are typically focused on flows on the two-dimensional plane, discretized with a structured grid (pixels of an image). The field is still active today, with a focus on inverse problems, uncertainty quantification and identification of partial differential equations~\cite{Sun-2018b}.

Here, we utilize the real-value non-volume preserving (RealNVP) methodology \cite{RealNVP} for normalizing flows which learns compositions of local affine transformations between the support of densities. In this methodology, the scaling and translation of the transformations over the support are parameterized as neural networks with an explicit dependence on physical time. This idea is similar to vector-valued optimal mass transport \cite{OMT2,OMT1} where a continuity PDE, analogous to hydrodynamic conservation laws, is utilized for computing the transport of densities. The distance between these densities is therefore minimized under the conservation constraints. A related work may be found in \cite{zou2005equation}, where equations for time-varying cumulative distribution functions could directly be learned given access to a fine-scaled simulator. Another relevant work in the literature was presented in \cite{hodgkinson2020stochastic} where SDEs are utilized to parameterize the transformation of a sample from a latent (generally a time-invariant multivariate isotropic Gaussian) to a time-varying target density. It is important to clarify here that concatenated transformations between supports can also be devised as flow maps governed by ordinary differential equations (ODEs) or SDEs (like in \cite{hodgkinson2020stochastic}). Note that the solutions of these equations are obtained through numerical discretization in \emph{fictitious} time and must not be confused with the physical time evolution of the {\em target} densities themselves; evolution in this fictitious time is then learned and deployed to construct each reference-to-target transformation.

The methodology we adopt in this article does not utilize the concept of an ODE- or SDE-solver in fictitious time; instead, we use compositions of stages of the previously mentioned affine transformations between supports. In such a manner, we are able to approximate densities from data without assumptions about the nature of the SDE or the corresponding Fokker-Planck equations.

To demonstrate the performance of our methodology, we take the solution of the Fokker-Planck equation for certain types of systems as a sequence of target distributions to be approximated. For this task, we first simulate sample paths from a stochastic differential equation to obtain our training data. At regular intervals in time, the positions of the various samples are measured from the sample paths.

This data is used to train the parameters of one physical-time-dependent normalizing flow which maps from the support of a reference (or latent) density to the density of the samples at different instances in physical time. For clarity, we note that the end result of this formulation is one set of neural networks that predict the transformation of the support, with physical time as an explicit input. Finally, the trained normalizing flow can be used to reproduce the evolving distribution and subsequently sample from it. This is achieved by noticing the fact that the reference density is a simple isotropic Gaussian (or a tractable initial condition of the Fokker-Planck equation evolution) that is easily sampled and then transformed to a sample from the target density at a specific time. We reiterate, to preclude any potential confusion, that the \emph{normalizing flow} we refer to, does not imply a flow in the sense of dynamical systems, but it embodies a bijective transformation, parameterized by time, from the support of a latent space distribution to that of the evolutionary distribution which is our target for learning. Similarly, we also emphasize, that the trained normalizing flow does not generate sample path data of the underlying process; instead it samples independently from several physical-time varying distributions. Finally, we note that the trained normalizing flow does not generate solutions to the Fokker-Planck equations with arbitrary initial and boundary conditions; it approximates the density for conditions that have generated the observed sample paths. We validate our proposed method for complex evolving target distributions  (e.g., SDEs with multiplicative noise and multimodal distributions).

The remainder of this article is arranged as follows: In Section 2, we briefly review normalizing flows and the previously proposed scalar temporally varying normalizing flow. In Section 3, we introduce our multivariate normalizing flow that learns time-varying distributions. In Section 4, we will show our results for different cases and compare with true density evolutions obtained from the Fokker-Planck equations as well as the results from its approximation through the Kramers–Moyal expansion \cite{KM,YangLi2020a,DaiMinChaos}.  Our final section contains a discussion of this study and perspectives for building on this line of research.

\section{Preliminaries}
\label{NF}

Normalizing flows provide a flexible and robust technique to sample from unknown probability distributions, through sampling a reference distribution and exploiting a series of bijective transformations from the support of the reference distribution to that of the target (unknown) density. In the following, we adopt the notation of \cite{NFs2}. Given a $D$-dimensional real vector $z\in \mathbb{R}^D$, sampled from a known reference density $p_{z}(z)$, the main idea of flow-based modeling is to find an invertible transformation $T$ such that:
\begin{align}
z=T(x), \quad \text{where} \quad z\thicksim p_{z}(z) \text{ and } x \thicksim p_{x} (x),
\end{align}
where $x \in \mathbb{R}^D$ and $p_x(x)$ is the unknown target density. When the transformation $T$ is invertible and both $T$ and $T^{-1}$ are differentiable, the density $p_x(x)$ can be obtained by a change of variables, i.e.,
\begin{align}\label{CoV}
p_{x}(x)=p_{z}(T(x))\mid {\rm det} J_{T}(x)\mid,
\end{align}
where $J_{T}(x)$ is the Jacobian of $T$. (Note that for cases where the transformation between densities
is not well captured by a simple transformation of the corresponding supports (e.g. when one of the densities is discontinuous), techniques in the spirit of attractor reconstruction for dynamical systems
have been proposed to extract a time-dependent normalizing flow  \cite{moosmuller2020geometric}). In the remainder of this article, we assume that a smooth and invertible transformation exists between reference and target distribution supports. Given two invertible and differentiable transformations $T_1$ and $T_2$, we have the following properties:
\begin{align}
(T_{2}\circ T_{1})^{-1} &= T_{1}^{-1}\circ T_{2}^{-1} \nonumber\\
{\rm det} J_{T_{2}\circ T_{1}(z)}&={\rm det} J_{T_{2}}(T_{1}(z))\cdot {\rm det} J_{T_{1}}(z). \nonumber
\end{align}
Consequently, we can build complex transformations by composing multiple simpler transformations, i.e., $T=T_{K}\circ T_{K-1}\circ \cdots \circ T_{1}$.

If we consider a set of training samples $\{x_1, \ldots, x_n \}$ taken from the target distribution $p_{x}(x)$, the negative log-likelihood given this data would be
\begin{align}\label{loss1}
\mathcal{L} = -\sum_{i=1}^n {\rm log} p_{x}(x_i).
\end{align}
Changing variables and substituting (\ref{CoV}) into (\ref{loss1}), we have
\begin{align}\label{loss2}
\mathcal{L} = -[\sum_{i=1}^n {\rm log} p_{z}(T(x_i)) + {\rm log}\mid {\rm det} J_{T}(x)\mid_{x=x_i}].
\end{align}
Therefore, in order to learn the transformation $T$, we only need to minimize the loss function given by Equation (\ref{loss2}). Here, we assume that the training set samples and the true data distribution sufficiently densely, so that maximizing the likelihood for the parameters of $T$ and then using $T^{-1}\left(\mathcal{N}(0,I)\right)$ approximates it well.

\section{Normalizing flows for time-varying distributions}

Normalizing flows have been successful for learning how to sample from stationary distributions.
A next step is to extend this sampling facilitation capability to distributions that vary in physical time.
An ad-hoc strategy for this involves embedding information about the timing of the samples into the training data, e.g., by appending a time stamp as an additional dimension of these data.
%%YGK
% Romit thus is the place to say it nicely
%
%%%YGk   The limitation of this approach is that when transformations are applied to this additional dimension, the invariant nature of physical time is lost when mapping from the support of latent to that of the target densities. To remedy this for scalar densities,
The authors in \cite{stNFs} proposed a way to extend scalar normalizing flows to the time-varying case, where successive snapshots of an evolving probability distribution $p_x(x,t)$ were mapped to {\em the same} reference distribution $p_z(z)$ through a time-dependent or, as they termed it,
``temporal" normalizing flow $x(t) = T^{-1}(z;t)$ \footnote{We explicitly define the physical time as a parameter of the transformation in contrast with the definitions in \cite{stNFs}}.  We will briefly review this method in the following, before we proceed to extend it to multidimensional evolving distributions.

Consider a time-varying probability density $p_{x}(x,t)$, where $x$ is a scalar random variable. If $z$ is the reference spatial coordinate, we would like to find an invertible and differentiable transformation $T$ such that $z=T(x;t), t\geq0$. This may be achieved by a transformation which possesses the following determinant of the Jacobian
\begin{center} {${\rm det}J_{T} $={\rm det}$\left[ {\begin{array}{*{30}{c}}
 \frac{\partial z}{\partial x} & \frac{\partial z}{\partial t}\\
 0 & 1
\end{array}} \right]$=$\frac{\partial z}{\partial x}$}
\end{center}

Therefore a normalizing flow between the reference and target density becomes,
\begin{align}
\label{tnf1}
p_{x}(x,t)=p_{z}(z,t)\mid \frac{\partial z}{\partial x}\mid, \quad where \quad z=T(x;t).
\end{align}
To be specific, the authors in \cite{stNFs} constructed a transformation by learning two functions
$f(x,t)$ and $z_0(t)$ so that
\begin{align}\label{tnf2}
z=\int e^{f(x,t)}dx +z_{0}(t),
\end{align}
where $z_{0}(t)$ is a time dependent offset function; the probability density then transforms as
\begin{align}
    p_{x}(x,t)=p_{z}(z,t)e^{f(x,t)}.
\end{align}
Both $f(x,t)$ and $z_{0}(t)$ were modeled as  unconstrained neural networks, trained via minimization of the negative log-likelihood function (\ref{loss2}). Note that the proposed transformation (equations (\ref{tnf1}) and (\ref{tnf2})) is not easily extendable to higher dimensions in its current form. Here, we propose an extension of the approach capable of learning time-varying transformations for \emph{multi-dimensional densities}.
We learn bijective mappings from the support of a reference probability density to that of a time-varying target density. For an effective transformation $T$, we must not only consider  invertibility and differentiability, but also the computational burden of the Jacobian matrix computation (thus Jacobian matrices which are diagonal are preferable). Inspired by \cite{RealNVP}, we propose a multivariate time-dependent RealNVP architecture to deal with multidimensional time-varying probability densities. If $p_{x}(x,t)$ is the target time-varying probability density, where $x=(x_1,\ldots,x_D)$ is a $D-$dimensional real vector and the reference density is denoted by $p_{z}(z)$, $z\in \mathbb{R}^D$, we aim to design an invertible and differentiable transformation with a computationally tractable Jacobian. Among many possible ways to construct bijective transformations using neural networks \cite{IAF,MAF,durkan2019neural} we propose here to extend the one-dimensional approach of \cite{stNFs} to compositions (possibly along permuted directions) of transformations of the following type
\begin{align}\label{T}
z_{1:d}&=x_{1:d}, \nonumber \\
z_{d+1:D}&=x_{d+1:D}\odot e^{\mu(x_{1:d},t)}+\nu(x_{1:d},t);
\end{align}
here the notation $\odot$ is the Hadamard product, $\mu$ and $\nu$ are two different neural networks from $\mathbb{R}^{d+1} \rightarrow \mathbb{R}^{D-d}$;  $d$ is a hyperparameter and $x_{1:d}=(x_1,x_2,\ldots,x_d)$. If we denote this transformation by $T$, the determinant of the Jacobian matrix becomes a computationally tractable expression given by
\begin{align}
    {\rm det}J_{T} = e^{\mu(x_{1:d},t)}.
\end{align}
Consequently, for a given dataset $\mathcal{D}=\{\mathcal{D}_j\}_{j=1}^{m}$, where $\mathcal{D}_j=\{x_{i}^j\}_{i=1}^{n}$ is sampled from $p_{x}(x,t_j)$, the negative log-likelihood becomes
\begin{align}\label{loss3}
\mathcal{L} &= -\sum_{j=1}^m\sum_{i=1}^n {\rm log} p_{x}(x_{i}^j,t_j) \nonumber \\
&=-[\sum_{j=1}^m\sum_{i=1}^n  {\rm log} p_{z}(T(x_{i}^j,t_j)) + {\rm log}\mid {\rm det} J_{T}(x,t)\mid_{x=x_{i}^j,t=t_j}],
\end{align}
where $m$ corresponds to the number of snapshots in time used for obtaining training samples and $n$ is the number of samples obtained for every $m$\textsuperscript{th} snapshot. Each sample path is obtained by sampling a random new initial condition and evolving it according to the SDE. By minimizing the loss function (Equation \ref{loss3}) to get the optimal parameters of neural networks $\mu$ and $\nu$, we learn a flexible, yet expressive transformation $T$ from the dataset $\mathcal{D}$. We can then use this transformation $x=T^{-1}(z,t)$ to sample from the time-varying distribution $p_x(x,t)$. This means that we have learned a generative machine learning model that constructs the solutions of the Fokker Planck equations from snapshots of ensembles of sample paths.

\section{Results}

\subsection{Approximating time-varying distributions from sample data}

\begin{figure}
\centering
\includegraphics[trim={0cm 0cm 0cm 0cm},clip,width=\textwidth]{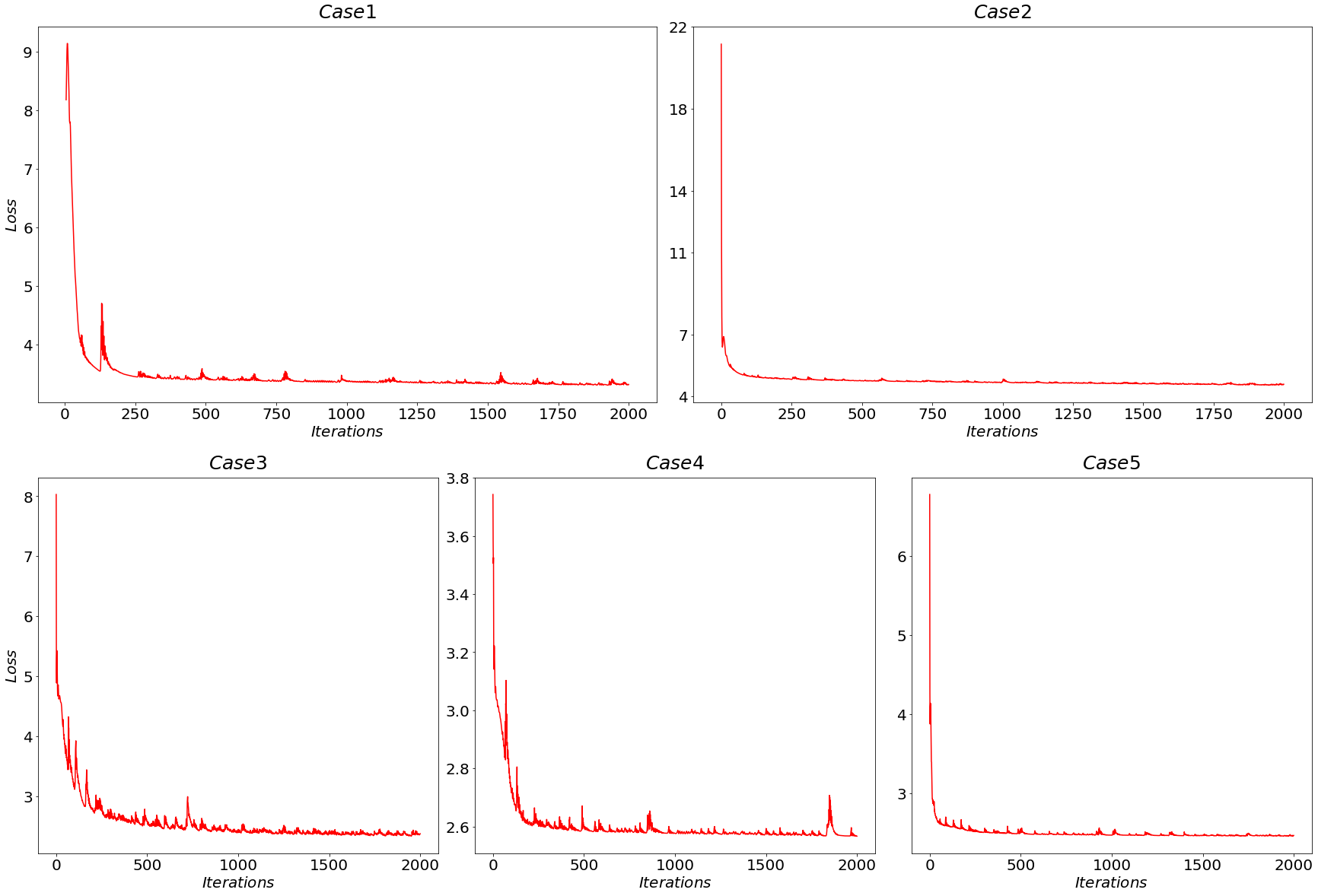}
\caption{Training progress to convergence for learning transformations to multivariate time-varying densities using normalizing flows; validation losses with increasing iterations are depicted. From left to right, these convergence histories correspond to the five cases that appear in our results.}
\label{FIGLoss}
\end{figure}

In the previous sections, we introduced a methodology to approximate (given appropriate samples) a time-varying family of multidimensional distributions through multivariate normalizing flows parameterized by physical time; at each moment in time, the corresponding normalizing flow maps our evolving multidimensional distribution back to the same, fixed, reference distribution. In this section, we provide illustrations of our method applied to data sampled from time-evolving multimodal distributions, such as those generated by evolving SDEs with multiplicative noise. In the following experiments, we set a learning rate of 0.005, our neural networks utilize 3 hidden layers, 32 neurons in each hidden layer, and the tan-sigmoidal activation for each hidden layer neuron. We use an Intel(R) Core(TM) I5-8265U CPU @1.60GHz for training of our neural networks. The convergence of the optimization is summarized in Figure \ref{FIGLoss}.

\begin{figure}
\centering
\includegraphics[trim={0cm 0cm 0cm 0cm},clip,width=\textwidth]{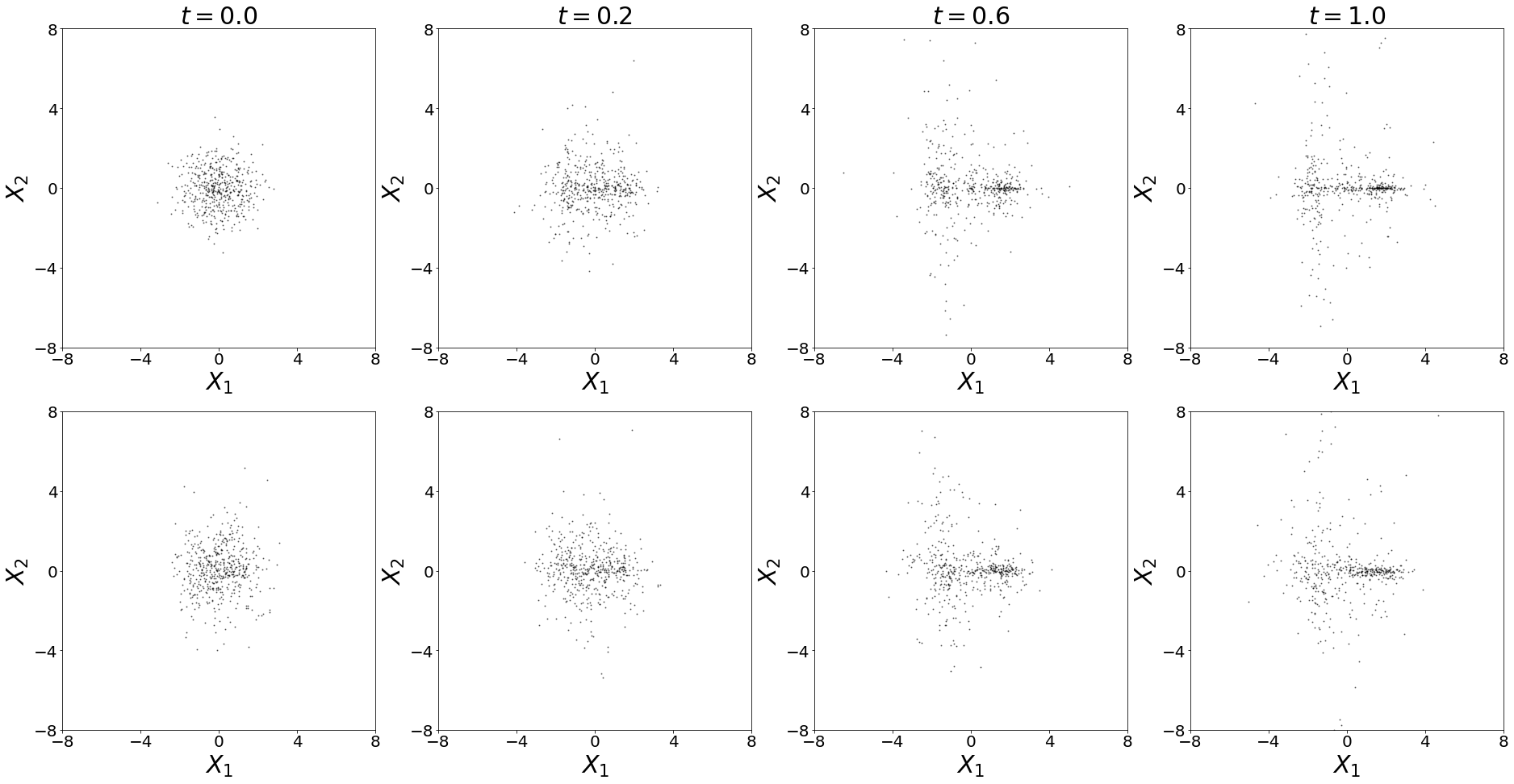}
\caption{Learning transformations from the support of an isotropic Gaussian distribution to the support of a 2D unimodal time-varying density using multivariate normalizing flows. Top row: Training data. Bottom row: samples obtained from learned distribution. We set the initial time $t_1=0$, the final time $t_m=1$, time step $\Delta t=0.05$, $m=21$ snapshots, and sample size $n=500$ as our dataset. Here we use $8$ compositions. Our neural networks utilize 3 hidden layers, 32 neurons in each hidden layer, and the tan-sigmoidal activation for each hidden layer neuron. The learning rate is 0.005.}
\label{FIGEx1Resampled}
\end{figure}

\begin{figure}
\centering
\includegraphics[trim={0cm 0cm 0cm 0cm},clip,width=\textwidth]{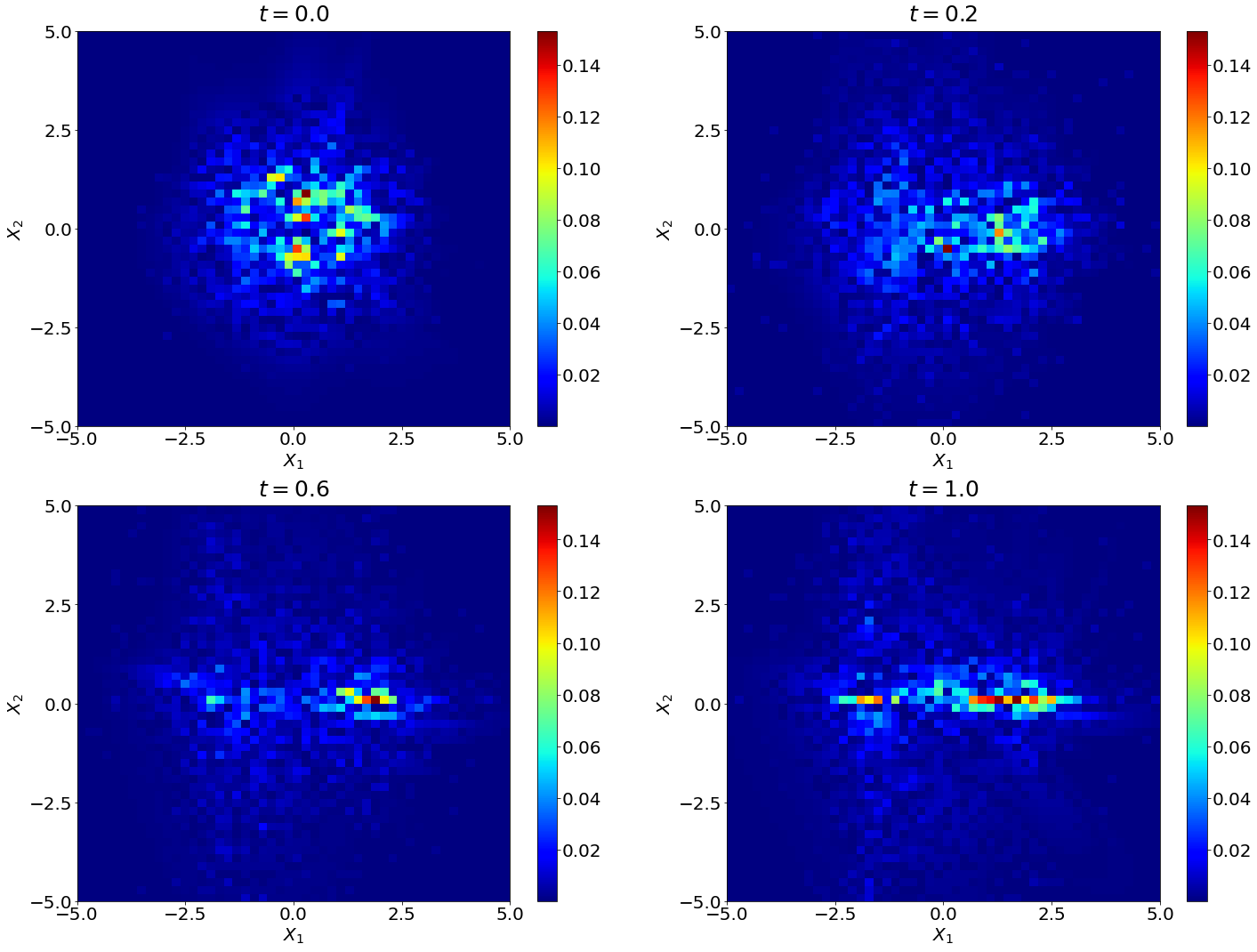}
\caption{Learning transformations from the support of an isotropic Gaussian distribution to the support of a 2D unimodal time-varying density using multivariate normalizing flows. This figure shows discrepancies between true and learned distribution, i.e., $|p_{true}-p_{learn}|$. }
\label{FIGEx1Accuracy}
\end{figure}

\noindent We first consider a two-dimensional SDE with pure jump L\'evy motion

given by
\begin{align}\label{EX1}
&dX_{1}(t)=(4X_{1}(t)-X_{1}^3(t))dt+X_{1}dL_{1}^{\alpha}(t), \nonumber \\
&dX_{2}(t)=-X_{1}(t)X_{2}(t)dt+X_{2}dL_{2}^{\alpha}(t),
%&(X_{1}(0), X_{2}(0))\thicksim \mathcal{N}(0, I_{2\times2}), \nonumber
\end{align}
where $L_{1}^{\alpha}$ and $L_{2}^{\alpha}$ are two independent, scalar, real-valued symmetric $\alpha-$stable L\'evy motions with triplet $(0,0,\nu_{\alpha})$. Here we take $\alpha=1.5$. The training data, denoted by $\mathcal{D}$, have been obtained by simulating the SDEs in Equation \eqref{EX1} with $n$ initial conditions sampled from $(X_{1}(0), X_{2}(0))\thicksim \mathcal{N}(0, I_{2\times2})$;  here $t_1=0$, $t_m=1$, $\Delta t=0.05$, $m=21$, and $n=500$.
We take the two-dimensional standard normal distribution as our reference density;
this also happens to be our initial condition, which then makes our time-dependent normalizing flow the time-T map of our FPE. In this two-dimensional problem, as described above in (\ref{T}), our normalizing flows consist of compositions of the following two types of prototypical transformations:
\begin{align}
\label{E1T1}
Y_{1}&=X_{1}, \nonumber \\
Y_{2}&=X_{2} e^{\mu(X_{1},t)}+\nu(X_{1},t),
\end{align}
and
\begin{align}
\label{E1T2}
Y_{1}&=X_{1} e^{\mu(X_{2},t)}+\nu(X_{2},t), \nonumber \\
Y_{2}&=X_{2},
\end{align}
to learn the map between the reference and the target supports. Note how the transformation in Equation \eqref{E1T2} performs a scaling and translation on the \emph{second} scalar component of the input vector, alternating from that performed in Equation \eqref{E1T1} on the {\em first} scalar component. Composition of such permuted affine transformations enable expressiveness in learning while preserving invertibility by construction. Here, we take 8 alternating compositions of transformations of the type given by Equations \eqref{E1T1} and \eqref{E1T2} and subsequently obtain $\mu_i, \nu_i$ with $i=1,2,...,8$ as the eight sets of neural networks that must be trained for this example. Each neural network also possesses an additional input neuron for the physical time. The final composition can also be expressed as follows:
\begin{align}\label{Transform}
\begin{gathered}
\begin{pmatrix}
1 & 0 \\ e^{\mu_1} & \nu_1
\end{pmatrix}
\rightarrow
\begin{pmatrix}
e^{\mu_2} & \nu_2\\
0 & 1
\end{pmatrix}
\rightarrow
\begin{pmatrix}
1 & 0 \\
e^{\mu_3} & \nu_3
\end{pmatrix}
\rightarrow
\begin{pmatrix}
e^{\mu_4} & \nu_4\\
0 & 1
\end{pmatrix}
\end{gathered}.
\end{align}
Note that our 16 neural networks ($\mu_i, \nu_i, i = 1,2,...,8$), must be trained jointly using  minimization of the negative log-likelihood function (\ref{loss3}). If the final composition is denoted by $T$, we can sample from the generated time-varying distribution, given knowledge of the reference density, and access to $T^{-1}$. We also compare samples from our training data with samples from the learned distribution facilitated through sampling the reference distribution and exploiting the learned transformation in Figure \ref{FIGEx1Resampled}.
%%%
We also compare the true density and the density generated through our time-varying normalizing flow in Figure \ref{FIGEx1Accuracy},
%%%YGK
\emph{both} for time instances used in training {\em and} for time instances
for which we had no training data.
%%%
Here we have used $t_1=0$, $t_m=1$, $\Delta t=0.05$, $m=21$ and $n=10^4$, sample paths to estimate each target density in a Monte Carlo procedure. For this problem we can actually numerically solve the FPE via a finite difference scheme; however, for higher dimensional cases, this finite difference discretization is generally a challenge \cite{TingGao,Xiaoli}.

Comparisons with numerical solutions for the associated FPE will be examined in following sections.

Now we consider a three-dimensional SDE with pure jump multiplicative L\'evy motion
\begin{align}\label{EX2}
&dX_{1}(t)=(4X_{1}(t)-X_{1}^3(t))dt+X_{1}dL_{1}^{\alpha}(t), \nonumber \\
&dX_{2}(t)=-X_{1}(t)X_{2}(t)dt+X_{2}dL_{2}^{\alpha}(t),\\
&dX_{3}(t)=X_{1}(t)X_{3}(t)dt+X_{3}dL_{3}^{\alpha}(t), \nonumber
\end{align}
where $L_{1}^{\alpha}$, $L_{2}^{\alpha}$ and $L_{3}^{\alpha}$ are three disjoint independent, scalar, real-valued, symmetric $\alpha-$stable L\'evy motions with triplet $(0,0,\nu_{\alpha})$. Here we take $\alpha=1.5$. The Fokker-Planck equation ofk this system is described in the Appendix in Equation (\ref{FPeqn}). We collect a dataset $\mathcal{D}$ from the numerical solution of this SDE, with initial conditions obtained from $(X_{1}(0), X_{2}(0), X_{3}(0))\thicksim \mathcal{N}(0, I_{3\times3})$, as our training data, where $t_1=0$, $t_m=1$, $\Delta t=0.05$, $m=21$ and $n=500$. We take the three-dimensional standard normal distribution as our reference (or latent) distribution (once again this corresponds to our Fokker-Planck initial condition). As in (\ref{T}), we learn the following six transformations and compose them (which are now parameterized by 12 neural networks).
\begin{align}
Y_{1}&=X_{1}, \nonumber \\
Y_{2}&=X_{2}, \\
Y_{3}&=X_{3}\odot e^{\mu(X_{1}, X_{2}, t)}+\nu(X_{1}, X_{2}, t), \nonumber
\end{align}

\begin{align}
Y_{1}&=X_{1}, \nonumber \\
Y_{2}&=X_{2}\odot e^{\mu(X_{1}, X_{3}, t)}+\nu(X_{1}, X_{3}, t),\\
Y_{3}&=X_{3}, \nonumber
\end{align}
and
\begin{align}
Y_{1}&=X_{1}\odot e^{\mu(X_{2}, X_{3}, t)}+\nu(X_{2}, X_{3}, t), \nonumber \\
Y_{2}&=X_{2}, \\
Y_{3}&=X_{3}. \nonumber
\end{align}
The corresponding progress to convergence of its optimization is also shown in Figure \ref{FIGLoss}. We compare the training data samples with samples from the reference distribution pushed forward to the target distributions at the desired times through the learned $T^{-1}$  in Figure \ref{FIGEx2Resampled}.

\begin{figure}[!htp]
\centering
\subfigure[X-Y plane]{
\label{XYplane}
\includegraphics[trim={0cm 0cm 0cm 0cm},clip,width=\textwidth]{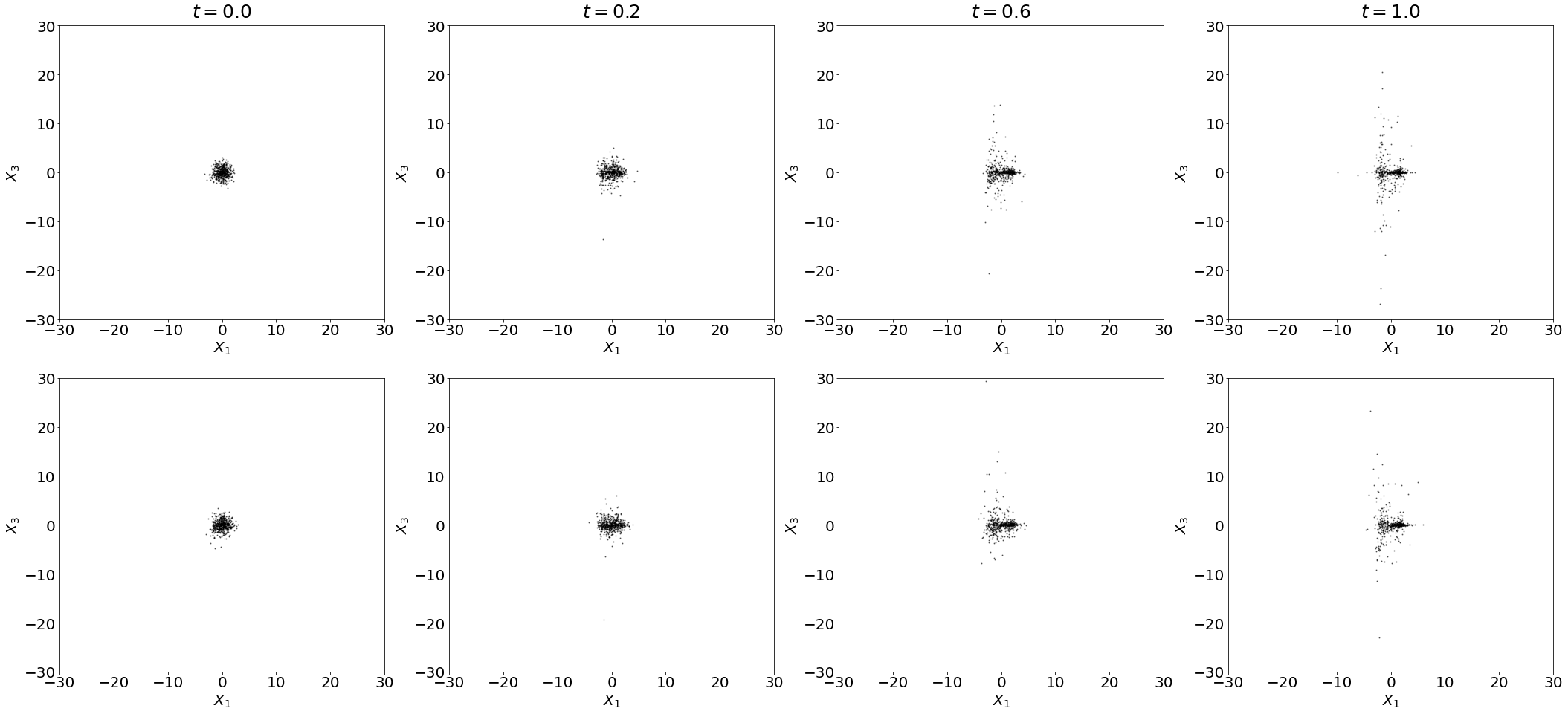}}

\subfigure[X-Z plane]{
\label{XZplane}
\includegraphics[trim={0cm 0cm 0cm 0cm},clip,width=\textwidth]{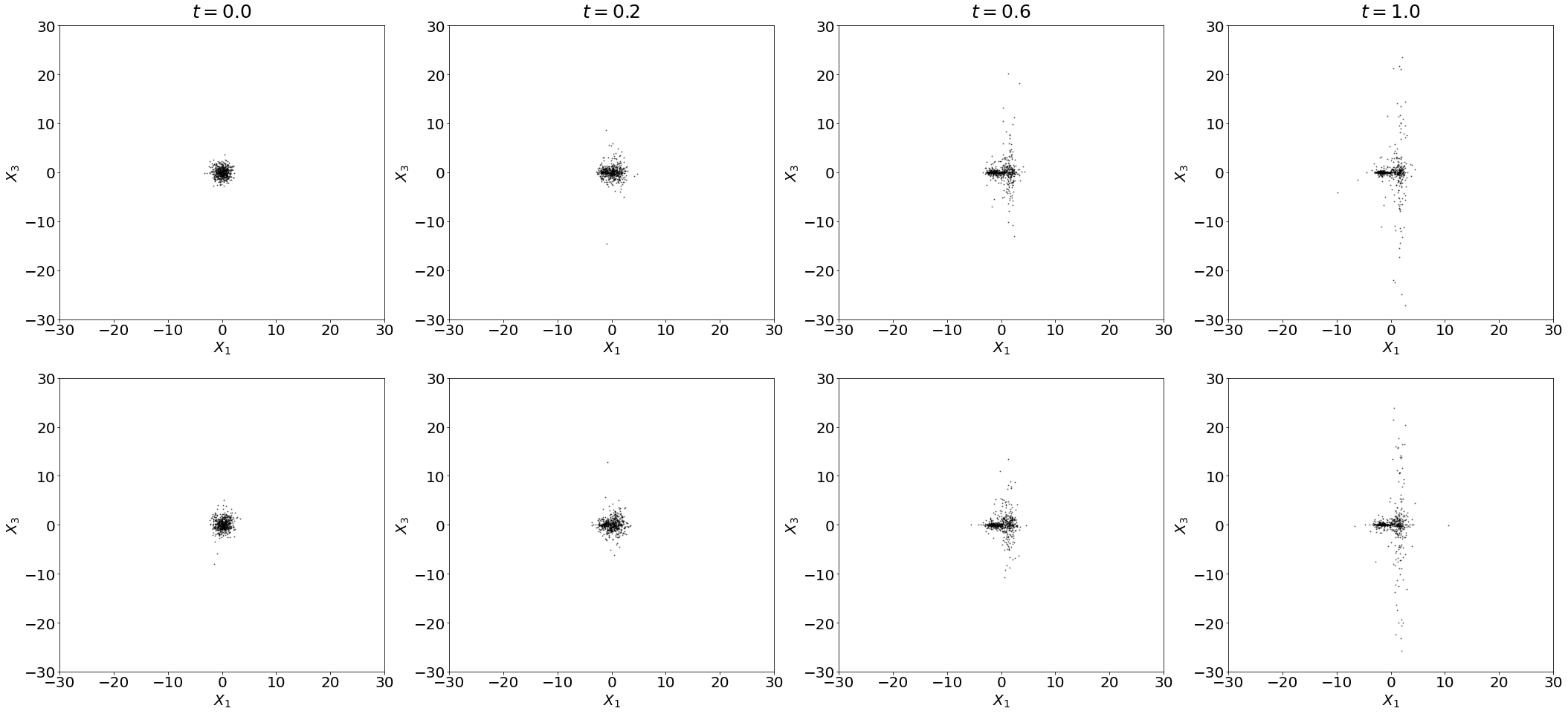}}
\label{FIGEx2Resampled}
\end{figure}

\begin{figure}[!htp]
\addtocounter{subfigure}{0}
\centering
% \ContinuedFloat
\subfigure[Y-Z plane]{
\label{YZplane}
\includegraphics[trim={0cm 0cm 0cm 0cm},clip,width=\textwidth]{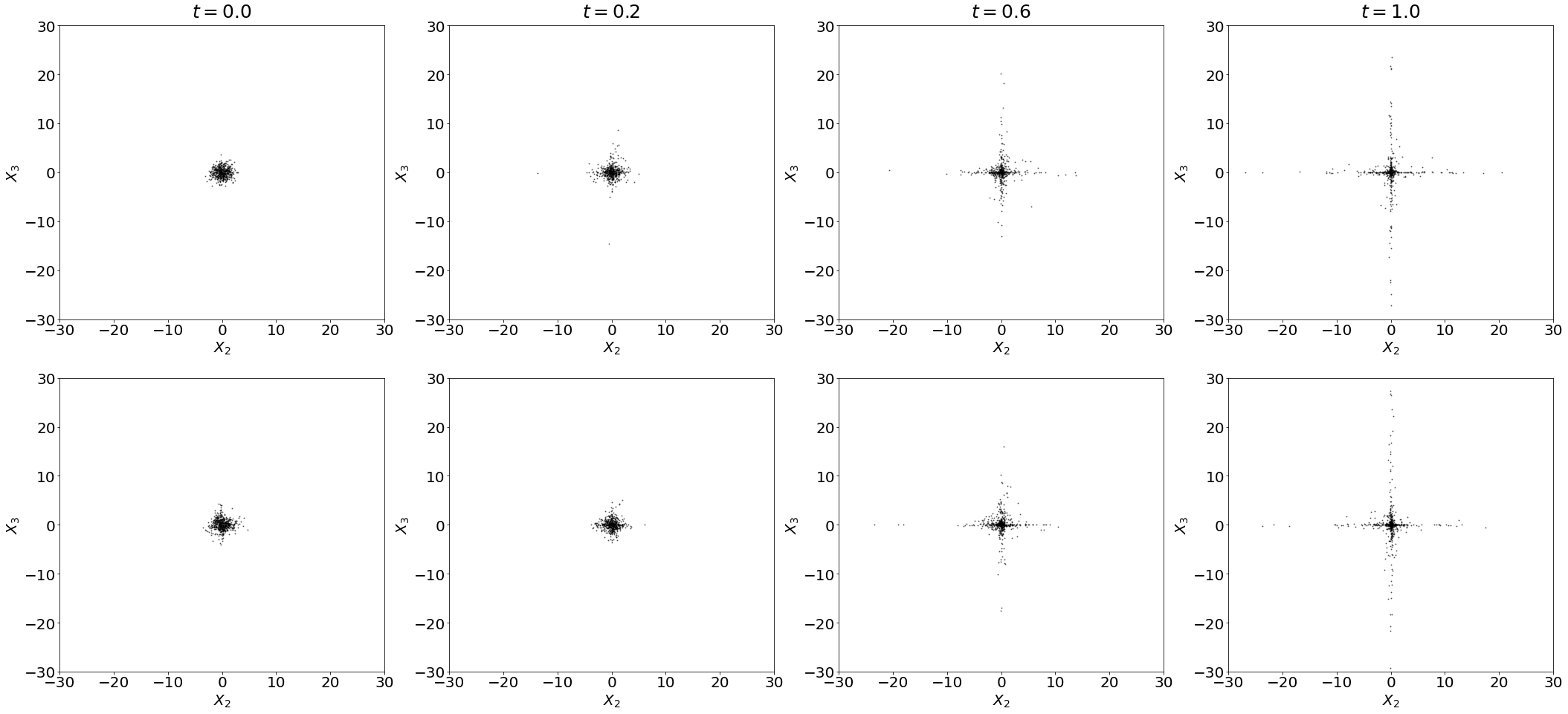}}
\caption{Learning 3D unimodal time-varying densities using multivariate normalizing flows. Top row: Training data. Bottom row: samples obtained from the generated distribution. (a) X-Y plane. (b) X-Z plane. (c) Y-Z plane. We set the initial time $t_1=0$, the final time $t_m=1$, time step $\Delta t=0.05$, $m=21$ snapshots, and sample size $n=500$ as our dataset. Here we use $6$ compositions. Our neural networks utilize 3 hidden layers, 32 neurons in each hidden layer, and the tan-sigmoidal activation for each hidden layer neuron. The learning rate is 0.005.}
\label{FIGEx2Resampled}
\end{figure}

\begin{figure}
\centering
\includegraphics[trim={0cm 0cm 0cm 0cm},clip,width=\textwidth]{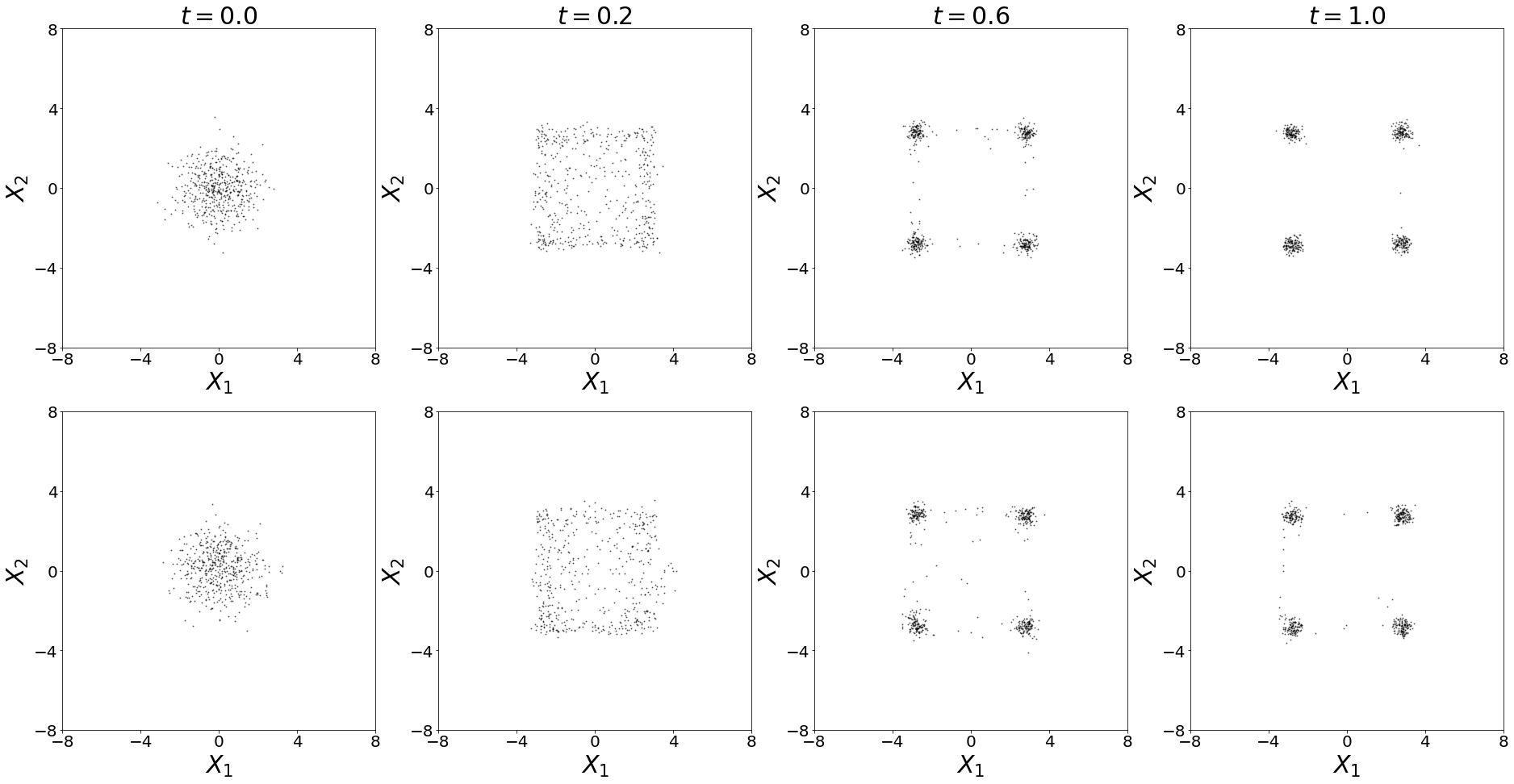}
\caption{Generating 2D multimodal time-varying densities using multivariate normalizing flows. Top row: Training data. Bottom row: samples obtained from generated distribution. We set the initial time $t_1=0$, the final time $t_m=1$, time step $\Delta t=0.05$, $m=21$ snapshots, and sample size $n=500$ as our dataset. Here we use $8$ compositions. Our neural networks utilize 3 hidden layers, 32 neurons in each hidden layer, and the tan-sigmoidal activation for each hidden layer neuron. The learning rate is 0.005.}
\label{FIGEx3Resampled}
\end{figure}

\begin{figure}
\centering
\includegraphics[trim={0cm 0cm 0cm 0cm},clip,width=\textwidth]{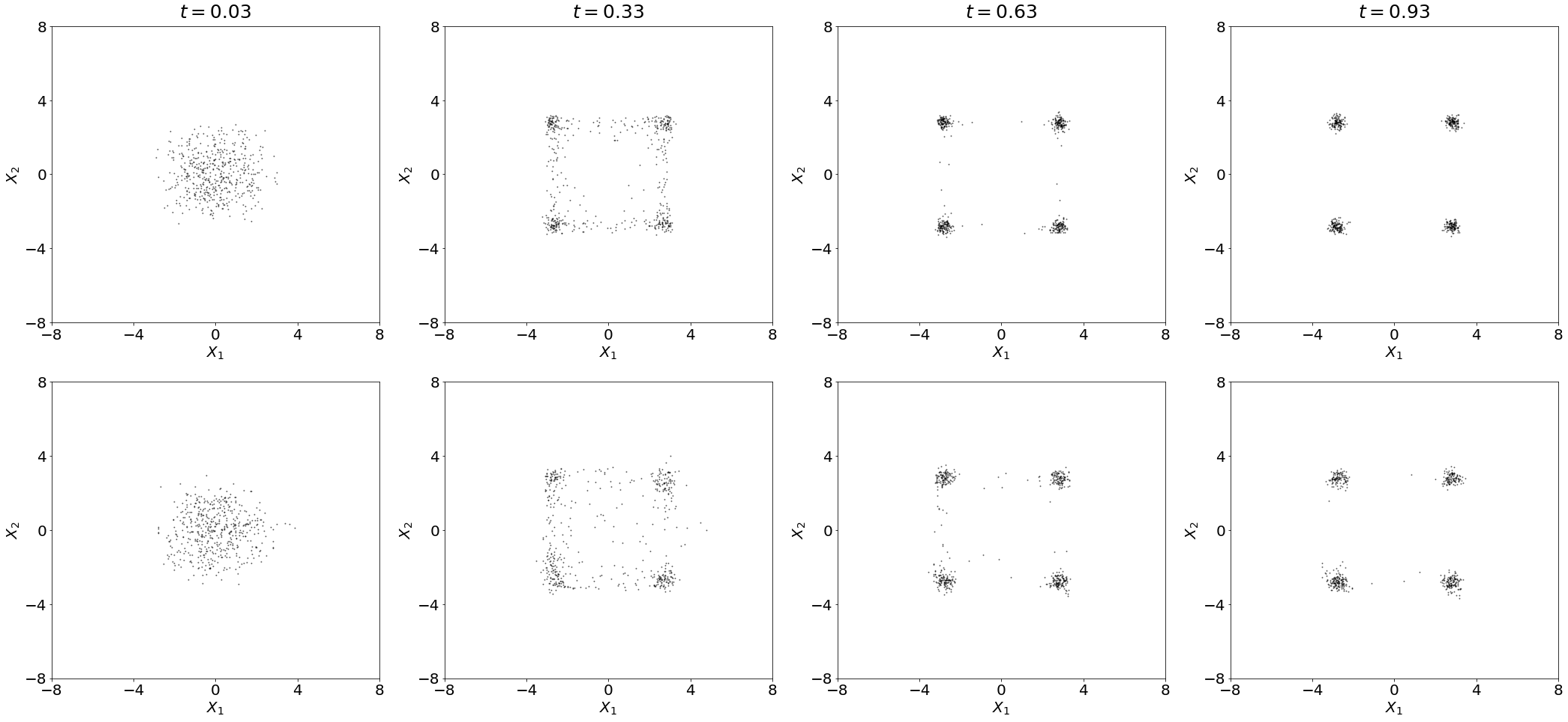}
\caption{Generating 2D multimodal time-varying densities using multivariate normalizing flows. Top row: test data. Bottom row: samples obtained from generated distribution.}
\label{FIGEx3Interpolating}
\end{figure}

After validating our method for 2D and 3D cases with unimodal time-varying distributions, we now consider a test case where a unimodal distribution evolves to a multimodal one with time. Given a two-dimensional SDE with Brownian motion
\begin{align}\label{EX3}
&dX_{1}(t)=(8X_{1}(t)-X_{1}^3(t))dt+dB_{1}(t), \nonumber \\
&dX_{2}(t)=(8X_{2}(t)-X_{2}^3(t))dt+dB_{2}(t),
\end{align}
where $B_{1}$ and $B_{2}$ are two independent scalar Brownian motions and initial conditions are given by $(X_{1}(0), X_{2}(0))\thicksim \mathcal{N}(0, I_{2\times2})$. Using a training data set that is generated using the same number of snapshots, and the same number of samples per snapshot as the example in Equation \eqref{EX1}, Figure \ref{FIGEx3Resampled} shows that we can capture the gradual evolution of a unimodal distribution into a multimodal one. Moreover, in Figure \ref{FIGEx3Interpolating}, we also validate the evolution in test data (In the training set, the time step $\Delta t=0.05$. Here we choose $t=0.03, 0.33, 0.63, 0.93$ as our test data). Note also, that we used the same architecture for the normalizing flow (i.e., given by Equation \eqref{Transform}).

\subsection{Comparisons with the Fokker-Planck equations}

We have demonstrated that our construction of time-dependent normalizing flows can learn transformations of a reference density to evolving, time-dependent ones by bijectively transforming the support of the reference density to that of the target densities. Although it is hard to obtain numerical solutions of the multidimensional Fokker-Planck equation, especially due to high dimensionality or non-locality (L\'evy noise case) in finite difference discretization, in this subsection, we will use simple examples to compare the probability densities generated from normalizing flows and those obtained from the associated Fokker-Planck equation.

Our first example will compare the solution obtained via our proposed method to that of the true Fokker-Planck equation. In addition, we use the Kramers-Moyal expansion \cite[Ch. 3]{KM} to identify the Fokker-Planck equation from data and compare our method to the solutions of this identified Fokker-Planck equation. Our system is given by a two-dimensional SDE with Brownian motion
\begin{align}\label{EX4}
&dX_{1}(t)=(4X_{1}(t)-X_{1}^3(t))dt+dB_{1}(t), \nonumber \\
&dX_{2}(t)=-X_{1}(t)X_{2}(t)dt+dB_{2}(t),
\end{align}
where $B_{1}$ and $B_{2}$ are two independent scalar Brownian motions. The initial condition of this system is given by a unimodal Gaussian density, i.e., $(X_{1}(0), X_{2}(0))\thicksim \mathcal{N}(0, \frac{1}{2}I_{2\times2})$. As shown below, this initial condition evolves into a multimodal density with time. For an SDE $dX_t=f(X_t) dt + \sigma(X_t) dB_t$ (which we remark, is driven by Brownian noise alone), the Kramers-Moyal expansions may be used to identify the drift $f(x)$ and diffusion $\sigma(x)$ from the sample path solutions of the SDE (see \cite{DaiMinChaos,YangLi2020a}),
\begin{align}\label{KMdrift}
f_{KM}(x)=\lim\limits_{\Delta t \to 0}\mathbb{E}[\frac{(X_{\Delta t}-X_0)}{\Delta t}|X_{0}=x],
\end{align}
\begin{align}\label{KMdiffusion}
\sigma_{KM}(x) \;\sigma_{KM}^T(x)=\lim\limits_{\Delta t \to 0}\mathbb{E}[\frac{(X_{\Delta t}-X_0)(X_{\Delta t}-X_0)^T }{\Delta t}|X_{0}=x].
\end{align}
These Kramers-Moyal formulas provide drift and diffusion coefficients which also appear in the so-called Kramers-Moyal expansion for the associated Fokker-Planck equation as in \cite[Ch. 3]{KM}. In other words, from the sample path data $X_t$, the Kramers-Moyal formulas offer us approximate
identifications of the corresponding SDE and FPE.
However, we note that this strategy for obtaining the coefficients is \emph{limited by the assumption of Brownian motion driven SDEs alone}. For the SDE given by Equation \eqref{EX4}, the corresponding Fokker-Planck equation is given by,
\begin{align}\label{FPeqn2}
p_t&=-\bigtriangledown\cdot(fp)+\frac{1}{2}{\rm Tr}[H(\sigma\sigma^{T}p)],
\end{align}
where the true solution is generated by $f(x)=[4x_1-x_{1}^3,-x_{1}x_2]^T$ and $\sigma=[1,0;0,1]$ and approximate solutions are obtained by using $f_{KM}(x), \sigma_{KM}(x)$ from the Kramers-Moyal expansion and through the use of normalizing flows. See Appendix for further details about the Fokker-Planck equation. This equation is subsequently solved (using both true and approximate drift and diffusion coefficients) using the finite-difference method for the purpose of comparison with the normalizing flow. Our spatial discretization uses a grid size of $\Delta x=\Delta y=0.2$, the time step size is $\Delta t=1.0e-4$ and the final time is given by $t_{m} = 1$.

% This scheme is also used to solve the \emph{true} Fokker-Planck equation constructed from the true drift and diffusion terms.

\begin{figure}
\centering
\includegraphics[trim={0cm 0cm 0cm 0cm},clip,width=\textwidth]{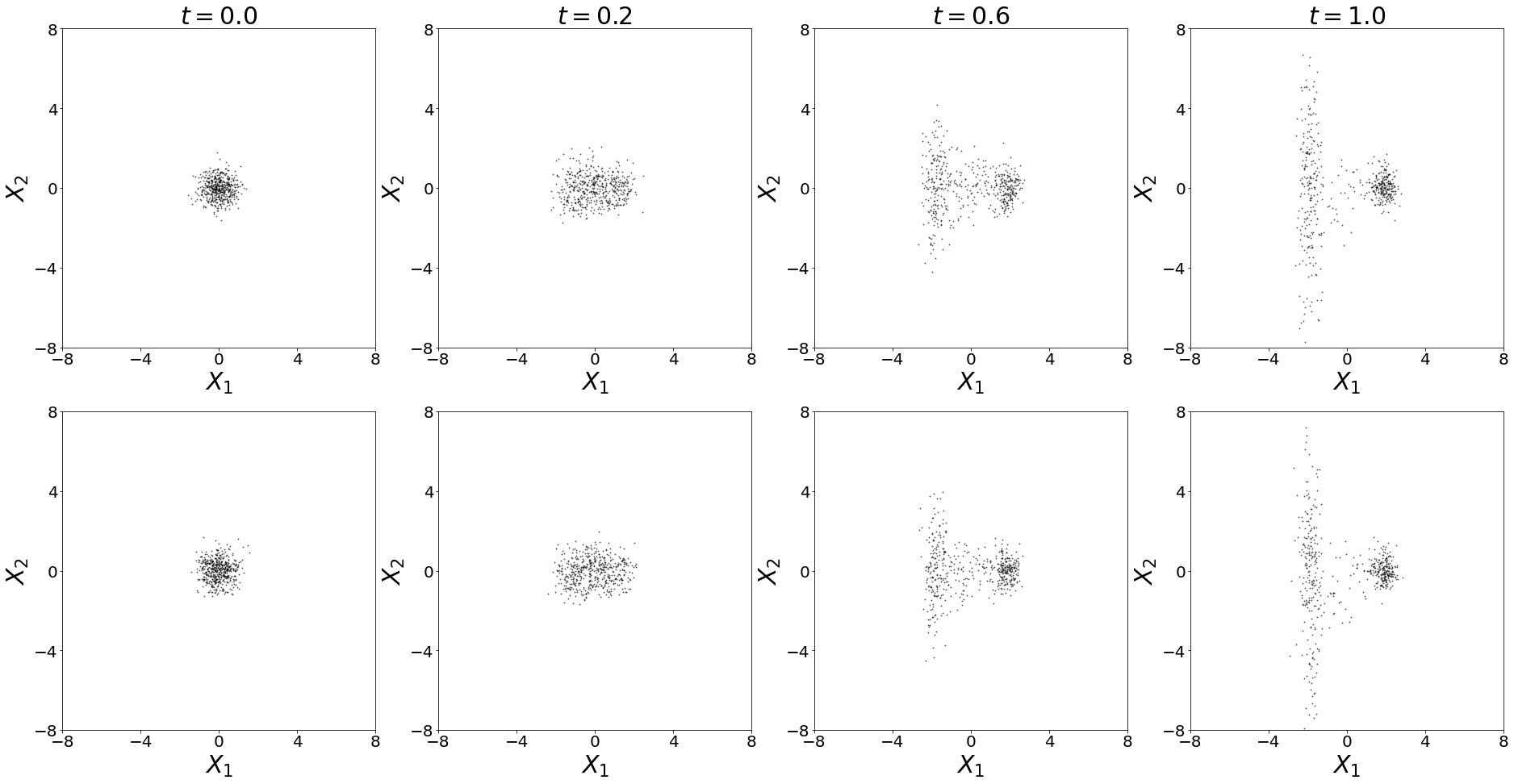}
\caption{Generating 2D unimodal time-varying densities using multivariate normalizing flows. Top row: Training data. Bottom row: samples obtained from generated distribution. We set the initial time $t_1=0$, the final time $t_m=1$, time step $\Delta t=0.05$, $m=21$ snapshots, and sample size $n=500$ as our dataset. Here we use $8$ compositions. Our neural networks utilize 3 hidden layers, 32 neurons in each hidden layer, and the tan-sigmoidal activation for each hidden layer neuron. The learning rate is 0.005.}
\label{FIGEx4Resampled}
\end{figure}

\begin{figure}
\centering
\includegraphics[width=\textwidth]{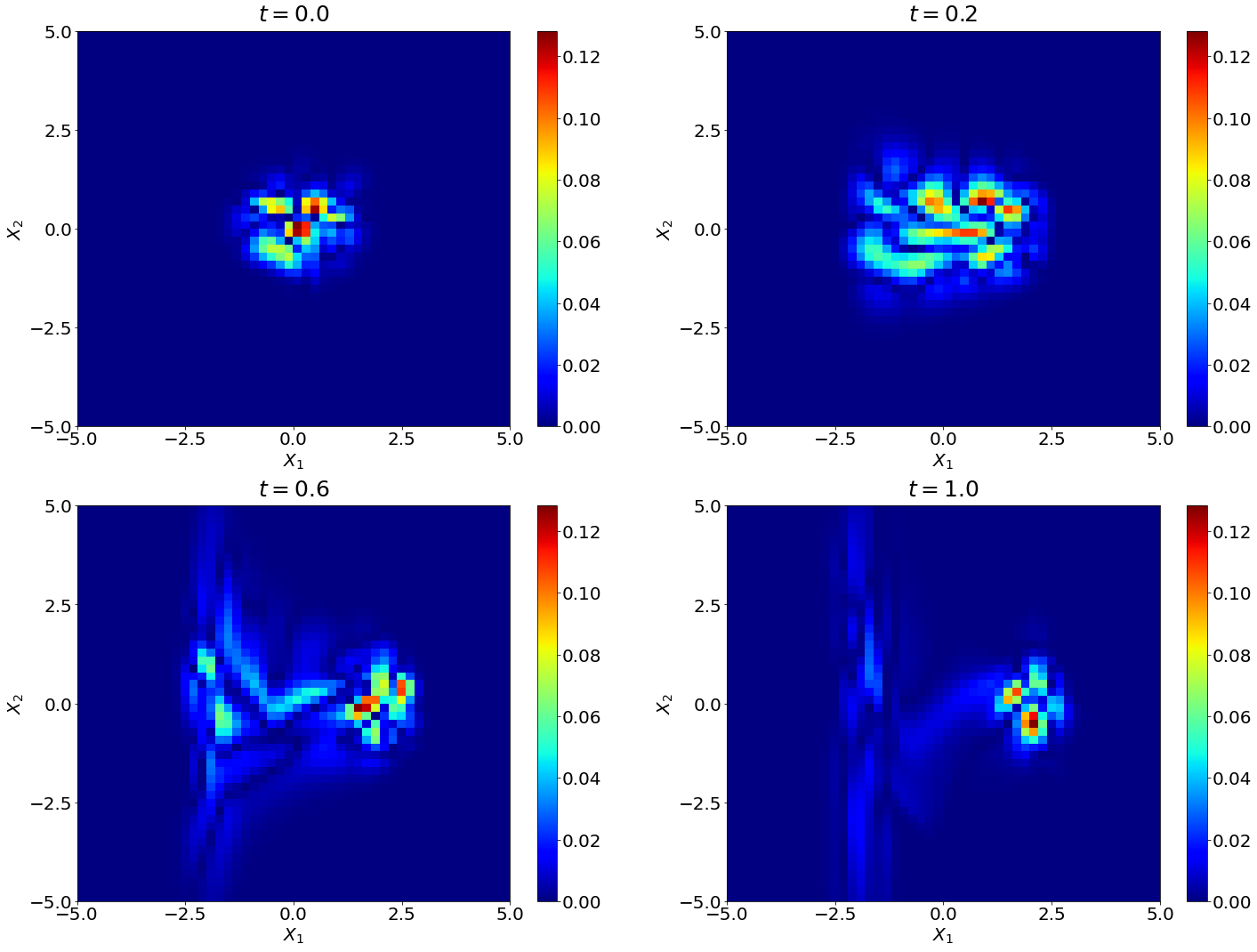}
\caption{Generating 2D unimodal density evolutions: The contours show  $|p_{true}-p_{learn}|$ to compare the solution of the Fokker-Planck equations and the generated density from the proposed method. Higher errors are observed in regions corresponding to higher probability mass. We attribute this to difficulties with learning a more complex mapping between source and latent supports of distributions.}
\label{FIGEx4Accuracy}
\end{figure}

\begin{figure}
\centering
\includegraphics[width=\textwidth]{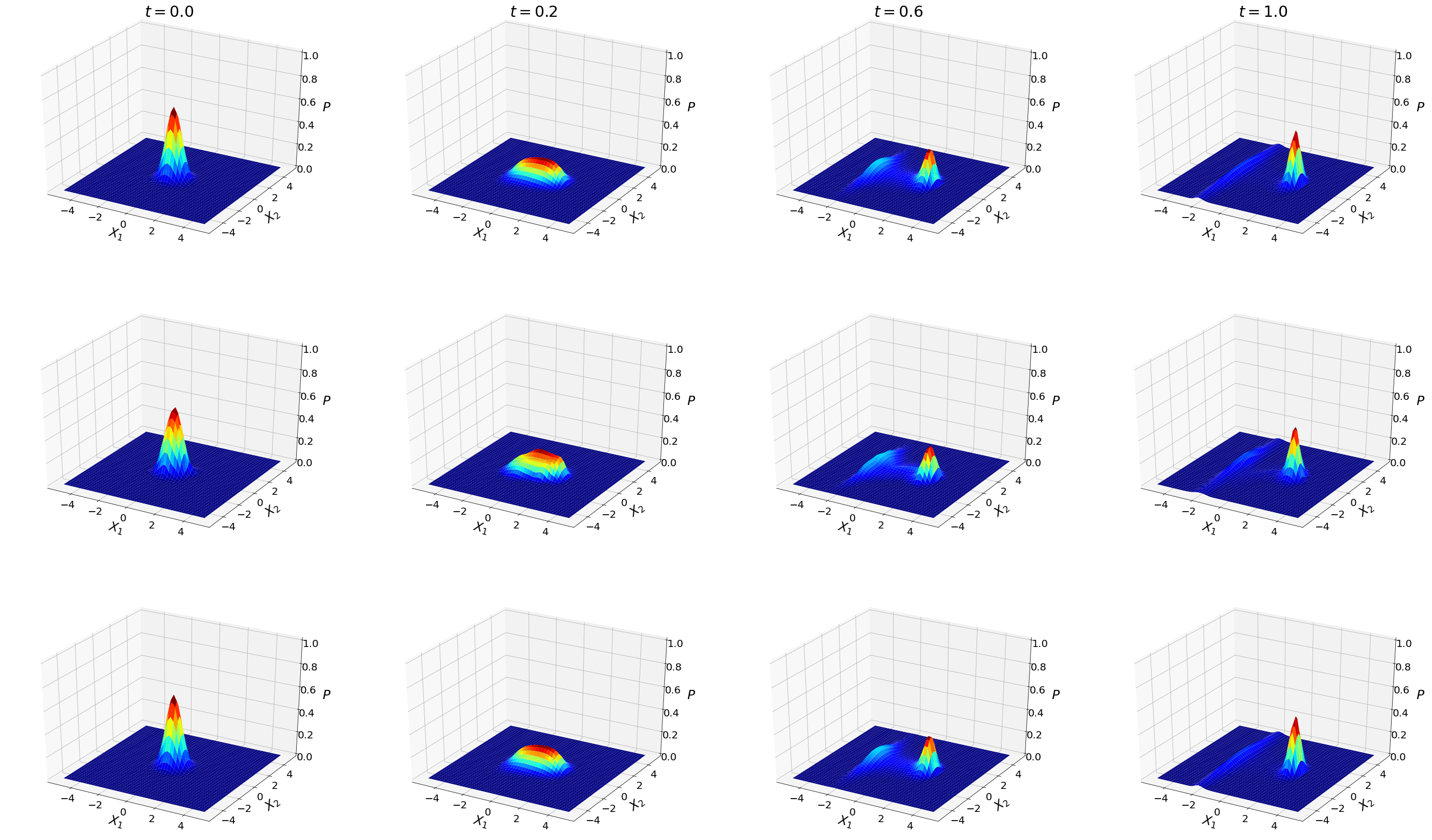}
\caption{Generating 2D unimodal time-varying densities. Top row: The true solution of Fokker-Planck equation. Middle row: The estimated probability density using the proposed method. Bottom row: The estimated probability density using the Kramers-Moyal expansion.}
\label{FIGEx4_KM}
\end{figure}

Figure \ref{FIGEx4Resampled} shows sample data from the training data set as well as from the generated distribution. It is readily observable that a multimodal structure emerges over time. Figure \ref{FIGEx4Accuracy}, containing $L_1$-norm errors, also shows that the estimated density is relatively accurate when compared to the true numerical solution of Fokker-Planck equation. It may be noticed here, that there are relatively larger errors in the regions of higher probability of the target density. Our assumption for the greater error arises from the fact that regions of higher probability in the target density may be associated with greater ``transport" from the latent Gaussian density and thus be prone to over- or undershoots after function approximation. Finally, Figure \ref{FIGEx4_KM} shows the density evolution of the true numerical solution of Fokker-Planck equation, one approximated by the normalizing flow and through the solution of the Fokker-Planck equation with drift and diffusivity obtained from the Kramers-Moyal formulas \eqref{KMdrift}-\eqref{KMdiffusion}. The results indicate that the proposed method is able to generate the correct density evolution. We remark that while identifying the FPE from data using the Kramers-Moyal formulas require assumptions about the nature of the noise; training a normalizing flow to learn the density directly does not require such assumptions.

%The corresponding Fokker-Planck equation as follow,
%\begin{align}\label{FPeqn3}
%p_t&=-\bigtriangledown\cdot(fp)+\frac{1}{2}{\rm Tr}[H(\sigma\sigma^{T}p)],
%\end{align}
%where $f(x)=[4x_1-x_{1}^3,-x_{1}x_{2}]^T$ and $\sigma=[1,0;0,1]$.

%% Add more detail about the non-local experiment - how was the Fokker-Planck obtained?

Our final representative example considers a two-dimensional SDE with pure jump L\'evy motion given by
\begin{align}\label{EX5}
&dX_{1}(t)=(X_{1}(t)-X_{1}^3(t))dt+dL_{1}^{\alpha}(t), \nonumber \\
&dX_{2}(t)=(X_{2}(t)-X_{2}^3(t))dt+dL_{2}^{\alpha}(t),
\end{align}
where $L_{1}^{\alpha}$ and $L_{2}^{\alpha}$ are two independent scalar $\alpha$-stable L\'evy motions with triplet $(0,0,\nu_{\alpha})$ and initial conditions $(X_{1}(0), X_{2}(0))\thicksim \mathcal{N}(0, \frac{1}{2}I_{2\times2})$. Here we take $\alpha=1.5$. The Fokker-Planck equation for the above SDE is given by
\begin{align}\label{FPeqn1}
p_t&=-\bigtriangledown\cdot(fp)+\int_{\mathbb{R}^2\backslash\{0\}} [p(x+\sigma y,t)-p(x,t)]\, \nu_{\alpha}(dy).
\end{align}

\begin{figure}
\centering
\includegraphics[trim={0cm 0cm 0cm 0cm},clip,width=\textwidth]{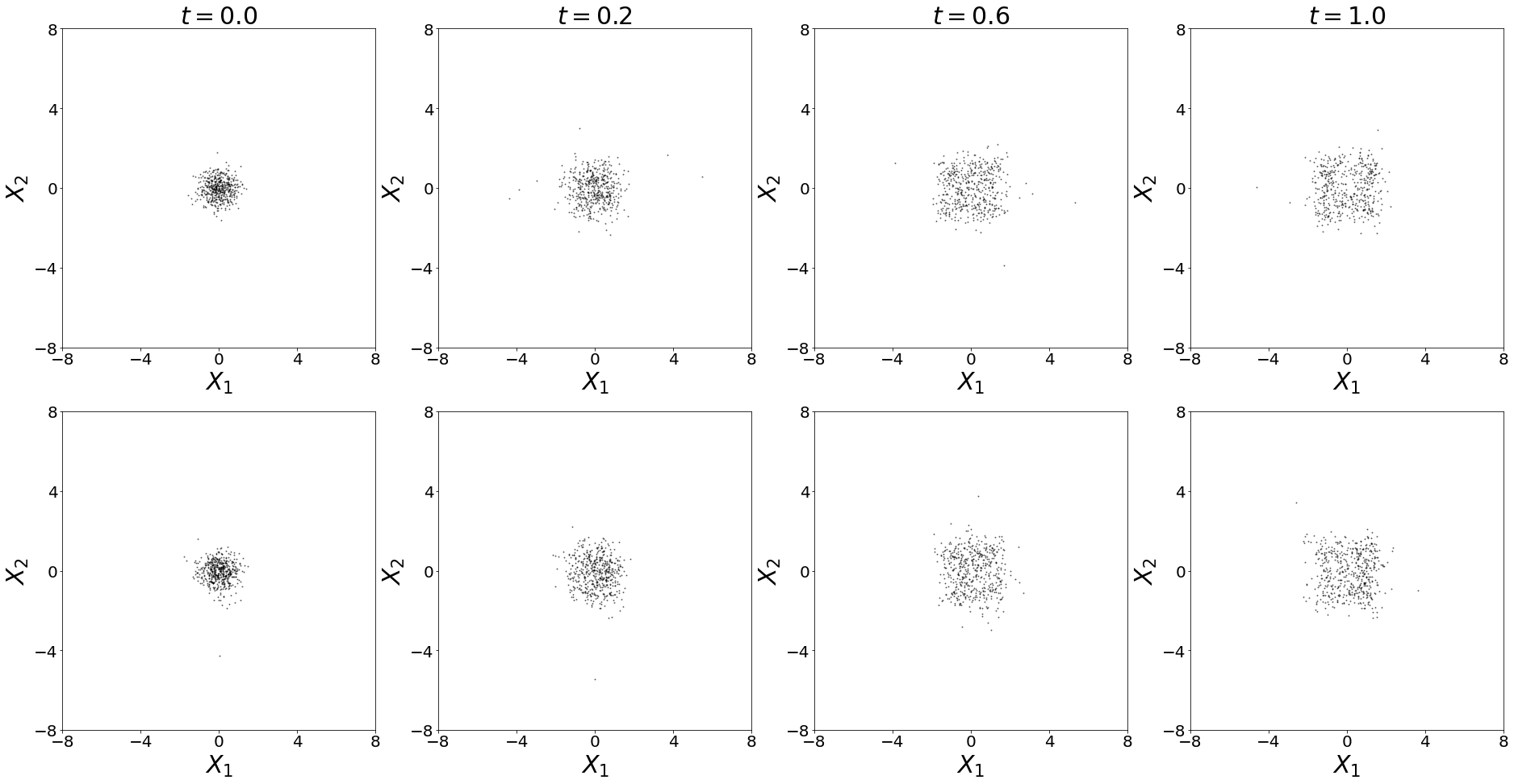}
\caption{Learning 2D multimodal time-varying densities using multivariate normalizing flows. Top row: Training data. Bottom row: samples obtained from learned distribution. We set the initial time $t_1=0$, the final time $t_m=1$, time step $\Delta t=0.05$, $m=21$ snapshotS, and sample size $n=500$ as our dataset. Here we use $8$ compositions. Our neural networks utilize 3 hidden layers, 32 neurons in each hidden layer, and the tan-sigmoidal activation for each hidden layer neuron. The learning rate is 0.005.}
\label{FIGEx5Resampled}
\end{figure}

\begin{figure}
\centering
\includegraphics[width=\textwidth]{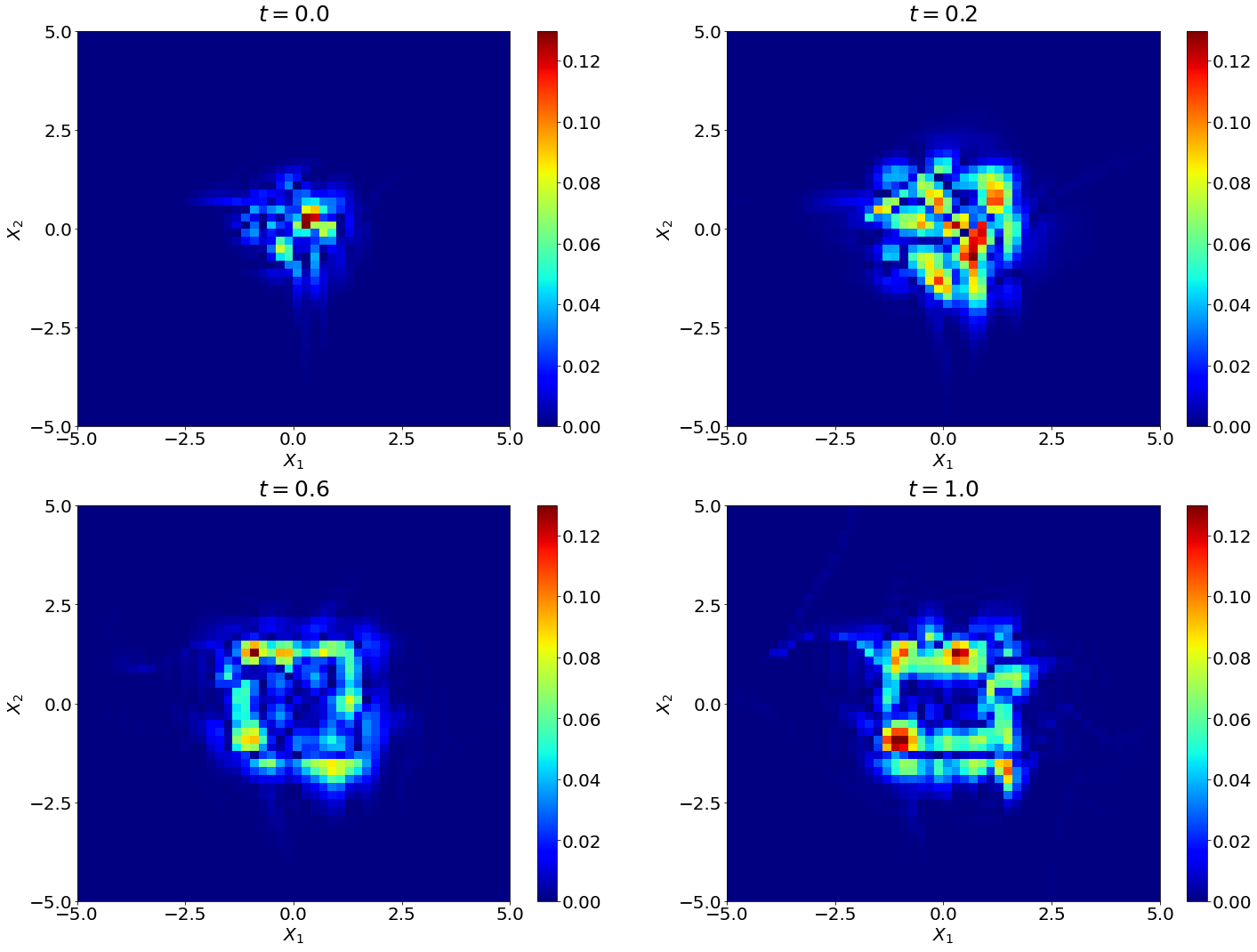}
\caption{2D multimodal density: Using formula $|p_{true}-p_{learn}|$ to compare the solution of Fokker-Planck equation and learned density.}
\label{FIGEx5Accuracy}
\end{figure}

\begin{figure}
\centering
\includegraphics[width=\textwidth]{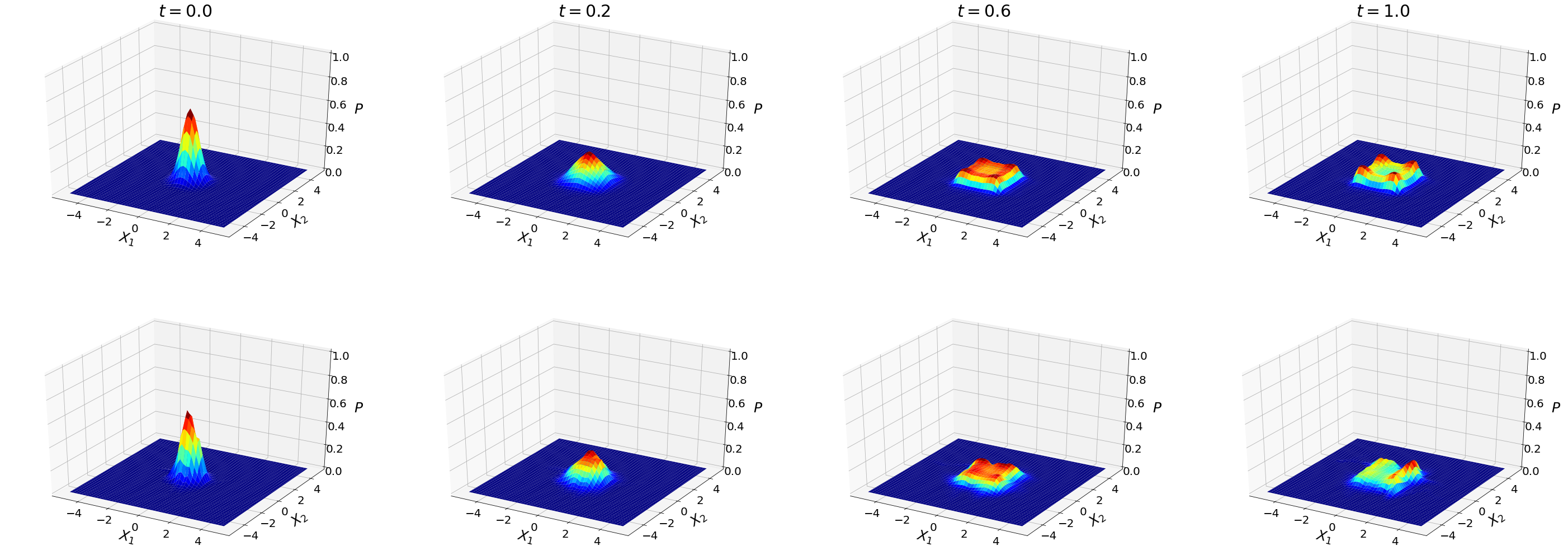}
\caption{Learning 2D multimodal time-varying densities. Top row: The true solution of Fokker-Planck equation. Bottom row: The estimated probability density using temporal normalizing flows.}
\label{FIGEx5_3d}
\end{figure}

\begin{figure}
\centering
\includegraphics[width=0.5\textwidth]{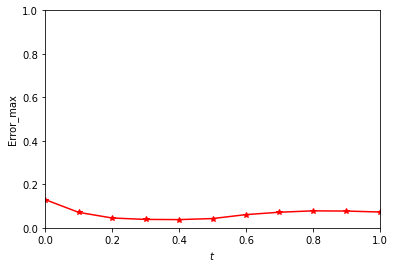}
\caption{The maximum error for different snapshots. }
\label{FIGEx5_error_max}
\end{figure}

\noindent
The jump measure is given by
\begin{align}
\nu_{\alpha}(dy)=c(2,\alpha)\parallel y\parallel^{-2-\alpha}dy, \nonumber
\end{align}
with $c(2,\alpha)=\frac{\alpha\Gamma((2+\alpha)/ 2)}{2^{1-\alpha}\pi\Gamma(1-\alpha/ 2)}$. See Appendix or \cite{Applebaum,Duan2015} for more details.\\
We note that this system of equations leads to a non-local FPE \eqref{FPeqn1} for which the Kramers-Moyal formulas  \eqref{KMdrift}-\eqref{KMdiffusion}, for drift and diffusivity, are invalid.The drift is $f(x)=[x_1-x_{1}^3,x_2-x_{2}^3]^T$, the diffusion is $\sigma=[1,0;0,1]$. In the absence of a Kramers-Moyal expansion for estimating the coefficients of the FPE, we directly solve the true nonlocal FP Equation (\eqref{FPeqn1}) via a finite difference scheme \cite{Xiaoli,TingGao}, for the purpose of comparison with the solution of the temporal normalizing flow. Our domain size is given by $[-10,10]\times[-10,10]$, with a spatial grid size of $\Delta x = \Delta y = 0.2$, a time step of $\Delta t = 1.0e-4$ and a final time of $t_m=1$. We emphasize, that in general we will not have access to a closed formula for this FPE, and that high dimensional equivalents are non-trivial to solve using the method of finite differences. Figure \ref{FIGEx5Resampled} shows that we can capture the gradual evolution of the distribution successfully in accordance with our previous examples. Figure \ref{FIGEx5Accuracy} also shows samples of both training data and those obtained from the generated densities using our temporal normalizing flow. Figure \ref{FIGEx5_3d} shows a qualitative comparison of the probability density evolution of the true numerical solution of the associated nonlocal Fokker-Planck equation (\ref{FPeqn1}) and that approximated by the temporal normalizing flows. Figure \ref{FIGEx5_error_max} shows the maximum error for different snapshots. It is seen that the data-driven density generation is successfully reconstructed.

\section{Discussion}

In this article, we have presented an approach for the machine learning assisted-estimation of 2- and 3-dimensional time-varying probability distributions from data generated by stochastic differential equations. Our generative machine learning technique, normalizing flows, rely on the estimation of a smooth invertible mapping from the support of a reference density (easily sampled from; in this study, a multivariate isotropic Gaussian), to that of an approximate target density evolving in physical time, consistent with our training data samples. This reference density may also simply chosen to be the initial condition of an evolving Fokker-Planck equation solution; then our normalizing flows embody the time-T map of the Fokker-Planck evolution. The mapping is estimated through  negative log-likelihood minimization and transforms the reference distribution support to that of the target time-dependent distribution(s). Once this support transformation is learned, samples from the reference density may be pushed-forward through it to obtain samples from the target time-varying density.

Our method extends previous work applicable solely for scalar time-varying stochastic processes,
and also demonstrates its applicability to non-local Fokker Planck dynamics arising, for example,
in modeling of stochastic systems involving L\'evy flights. We validate our method through numerical experiments on the time-varying densities of two- and three-dimensional stochastic processes. Our results compare favorably with the true solutions to the corresponding Fokker-Planck equations (when available) and to approximations of these Fokker-Plancks through estimates of the drifts and diffusivities of the associated Langevin SDEs using the Kramers-Moyal formulas. Based on these initial promising results from these experiments, we plan to explore higher dimensional applications and the concommitant numerical complexities. To that end, our plan is to investigate intelligent multiscale sampling strategies and alternative expressions (beyond the composition of affine transforms used here) for the invertible mappings between the supports of the reference and target densities; these alternatives may retain the scaling and translation transformations but alter the neural network architecture for them - or they may completely move away from these
affine transformations to more general ones. One may even contemplate extending to
noninvertible support transformations, in the spirit of \cite{moosmuller2020geometric}.

More importantly, given a learned  bijective transformation between supports, the approach may be utilized for the convenient extraction of the infinitesimal generator (i.e., the derivative) of the density evolution by automatic differentiation. By virtue of access to such a generator, we may use established as well as novel tools for data-driven identification of partial differential equation laws \cite{Kevrekidis4} to learn the Fokker-Planck equation \emph{for arbitrary initial conditions}, or the support evolution equation directly. This is the focus of current and future investigations.

%our next steps are to demonstrate the feasibility of these approaches for high dimensional problems and for problems where distributions may have further complexities in these high dimensions.

%Possible avenues for exploration also include the relaxation of the Markovian assumption for the invertible map construction (note how the present mapping from reference to target supports depend one one particular instance in time alone), with transformation from latent to target space being conditioned on time-delay embeddings now requiring information about path trajectories of samples.

% we can learn the infinitesimale generator (the derivative) of the density (like the FP, by automatic differentiation)
% and (NOT IN ABSTRACT) we can eveb learn the velocity field.

% We have presented an approach for estimating high-dimensional time-varying probability distributions, i.e., the solution of Fokker-Planck equation here (2D and 3D). We also validated that our method is robust for complex distribution, such as multimodal distribution. \\

% However,  there are several challenges for further investigation. First, reduced computation for higher dimensional problems(e.g., 10D problems). Second, improve the accuracy of density estimation especially for complex distribution.\\

\section*{Acknowledgments}
We would like to thank Xiaoli Chen, William McClure  and Stanley Nicholson for helpful discussions. This material is based upon work supported by the U.S. Department of Energy (DOE), Office of Science, Office of Advanced Scientific Computing Research, under Contract DE-AC02-06CH11357. This research was funded in part and used resources of the Argonne Leadership Computing Facility, which is a DOE Office of Science User Facility supported under Contract DE-AC02-06CH11357. R.~M. acknowledges support from the Margaret Butler Fellowship at the Argonne Leadership Computing Facility. The work of I.G.K was partially supported by the US DOE and by an ARO MURI.

\section*{Data Availability}
 Research code is shared via \url{https://github.com/Yubin-Lu/Temporal-normalizing-flows-for-SDEs/tree/main/TNFwithRealNVP}.

%%%%%%%%%%%  References

%\bibliographystyle{abbrv}
%\bibliography{references}
\bibliographystyle{unsrt}

\section*{Appendix}\label{App}
\setcounter{equation}{0}
\setcounter{subsection}{0}
\renewcommand{\theequation}{A.\arabic{equation}}
\renewcommand{\thesubsection}{A.\arabic{subsection}}

\subsection*{L\'evy processes}
%\textbf{L\'evy motions} \\
Let $L=(L_{t},t\geq0)$ be a stochastic process defined on a probability space $(\Omega, \mathcal{F}, P)$. We say that L is a L\'evy motion if:
\begin{itemize}
\item $L_{0}=0 (a.s.)$;
\item $L$ has independent and stationary increments;
\item $L$ is stochastically continuous, i.e., for all $a>0$ and for all $s\geq0$
$$
\lim\limits_{t\to s}P(|L_{t}-L_{s}|>a)=0.
$$
\end{itemize}

\subsection*{Characteristic functions}
%\textbf{Characteristic functions}\\
For a L\'evy motion $(L_{t},t\geq0)$, we have the L\'evy-Khinchine formula given by,
$$
\mathbb{E}[e^{i(u,L_t)}] = exp\{t[i(b,u)-\frac{1}{2}(u,Au)+\int_{\mathbb{R}^d\backslash\{0\}} [e^{i(u,y)}-1-i(u,y)I_{\{\Vert y\Vert<1\}}(y)]\, \nu(dy)]\},
$$
for each $t\geq0$, $u\in\mathbb{R}^n$, where $(b,A,\nu)$ is the triple of L\'evy motion $(L_{t}, t\geq0)$. \\

\noindent \textbf{The L\'evy-It\^{o} decomposition theorem:} If $(L_{t},t\geq0)$ is a L\'evy motion with $(b,A,\nu)$, then there exists $b\in\mathbb{R}^n$, a Brownian motion $B_{A}$ with covariance matrix $A$ and an independent Poisson random measure $N$ on $\mathbb{R}^{+}\times(\mathbb{R}^n-\{0\})$ such that, for each $t\geq0$, \\
$$
L_{t}=bt+B_{A}(t)+\int_{0<|x|<c} x\, \tilde{N}(t,dx)+\int_{|x|\geq c} x\, N(t,dx),
$$
where $\int_{|x|\geq c} x\, N(t,dx)$ is a Poisson integral and $\int_{0<|x|<c} x\, \tilde{N}(t,dx)$ is a compensated Poisson integral defined by
$$
\int_{0<|x|<c} x\, \tilde{N}(t,dx)=\int_{0<|x|<c} x\, N(t,dx)-t\int_{0<|x|<c} x\, \nu(dx).
$$
\\
\subsection*{The Fokker-Planck equations}
Consider a stochastic differential equation in $\mathbb{R}^n$:
\begin{align}\label{sde}
dX_t=f(X_{t-})dt+\sigma(X_{t-})dB_t + dL_{t}^{\alpha},\quad X_0=x_0,
\end{align}
where $f$ is a vector field, $\sigma$ is a $n\times n$ matrix, $B_t$ is Brownian motion in $\mathbb{R^n}$ and $L_{t}^{\alpha}$ is a symmetric $\alpha-$stable L\'evy motion in $\mathbb{R}^n$ with the generating triplet $(0, 0, \nu_{\alpha})$. The jump measure
\begin{align}
\nu_{\alpha}(dy)=c(n,\alpha)\parallel y\parallel^{-n-\alpha}dy, \nonumber
\end{align}
with $c(n,\alpha)=\frac{\alpha\Gamma((n+\alpha)/ 2)}{2^{1-\alpha}\pi^{n/2}\Gamma(1-\alpha/ 2)}$. The processes $B_t$ and $L_t^\alpha$ are taken to be independent.\\
The Fokker-Planck equation for the stochastic differential equation (\ref{sde}) is then given by
\begin{align}\label{FPeqn}
p_t&=-\bigtriangledown\cdot(fp)+\frac{1}{2}{\rm Tr}[H(\sigma\sigma^{T}p)]\\
&+\int_{\mathbb{R}^n\backslash\{0\}} [p(x+y,t)-p(x,t)]\, \nu_{\alpha}(dy),
\end{align}
where $p(x,0)=\delta(x-x_0)$. See \cite{Applebaum,Duan2015} for more details.

\end{document}